\documentclass[10pt,twocolumn,letterpaper]{article}

\usepackage{cvpr}
\usepackage{times}
\usepackage{epsfig}
\usepackage{graphicx}
\usepackage{amsmath}
\usepackage{amssymb}

\usepackage{algorithm}
\usepackage{algorithmic}
\usepackage{caption}
\usepackage{subcaption}

\usepackage[breaklinks=true,bookmarks=false]{hyperref}

\cvprfinalcopy 

\def\cvprPaperID{****} 

\ifcvprfinal\fi
\begin{document}

\title{Adversarial Representation Active Learning}

\author{Ali Mottaghi\\
Stanford University\\
{\tt\small mottaghi@stanford.edu}
\and
Serena Yeung\\
Stanford University\\
{\tt\small syyeung@stanford.edu}
}

\maketitle
\thispagestyle{empty}

\begin{abstract}
   Active learning aims to develop label-efficient algorithms by querying the most informative samples to be labeled by an oracle. The design of efficient training methods that require fewer labels is an important research direction that allows more effective use of computational and human resources for labeling and training deep neural networks. In this work, we demonstrate how we can use recent advances in deep generative models, to outperform the state-of-the-art in achieving the highest classification accuracy using as few labels as possible. Unlike previous approaches, our approach uses not only labeled images to train the classifier but also unlabeled images and generated images for co-training the whole model. Our experiments show that the proposed method significantly outperforms existing approaches in active learning on a wide range of datasets (MNIST, CIFAR-10, SVHN, CelebA, and ImageNet).  
   \footnote{Our code is available at  \url{https://github.com/samottaghi/adversarial-representation-active-learning}.}
\end{abstract}

\section{Introduction}

While deep learning has achieved great success in computer vision tasks ranging from image classification to detection and segmentation, these successes continue to require large amounts of labeled training examples. This is a significant challenge as deep learning practitioners increasingly apply deep learning models to solve new problems in diverse domains ranging from medicine \cite{yeung2019computer} to sustainability \cite{jean2016combining}, placing a suboptimal and sometimes prohibitive burden on domain experts to label large amounts of training data. Active learning, where training examples are incrementally selected for labeling to yield high classification accuracy at low labeling budgets, has therefore emerged as an exciting paradigm with significant potential for democratizing the use of deep learning \cite{gal2017deep}.

Many of the most successful approaches for active learning thus far have been based on pool-based active learning \cite{settles2009active} \cite{gal2017deep} \cite{sinha2019variational}, where small subsets of examples from a large unlabeled pool are iteratively selected for labeling, based on an acquisition function that assesses how informative the subset is expected to be for the training process. As the selected subsets are labeled by an oracle (i.e., a human annotator), they are added to a labeled dataset and used to update a classifier trained on the dataset. Much work in active learning has focused on developing effective acquisition functions, including those which select examples that produce high classifier uncertainty \cite{mackay1992information}, that are expected to lead to a high improvement in a Bayesian framework \cite{gal2017deep}, or that are not well represented in the labeled set \cite{settles2009active}. However, all of these approaches, at small labeling budgets, still suffer from a significant gap in performance when compared with training on large labeled datasets \cite{sener2017active}.

\begin{figure}[t]
\begin{center}
  \includegraphics[width=1\linewidth]{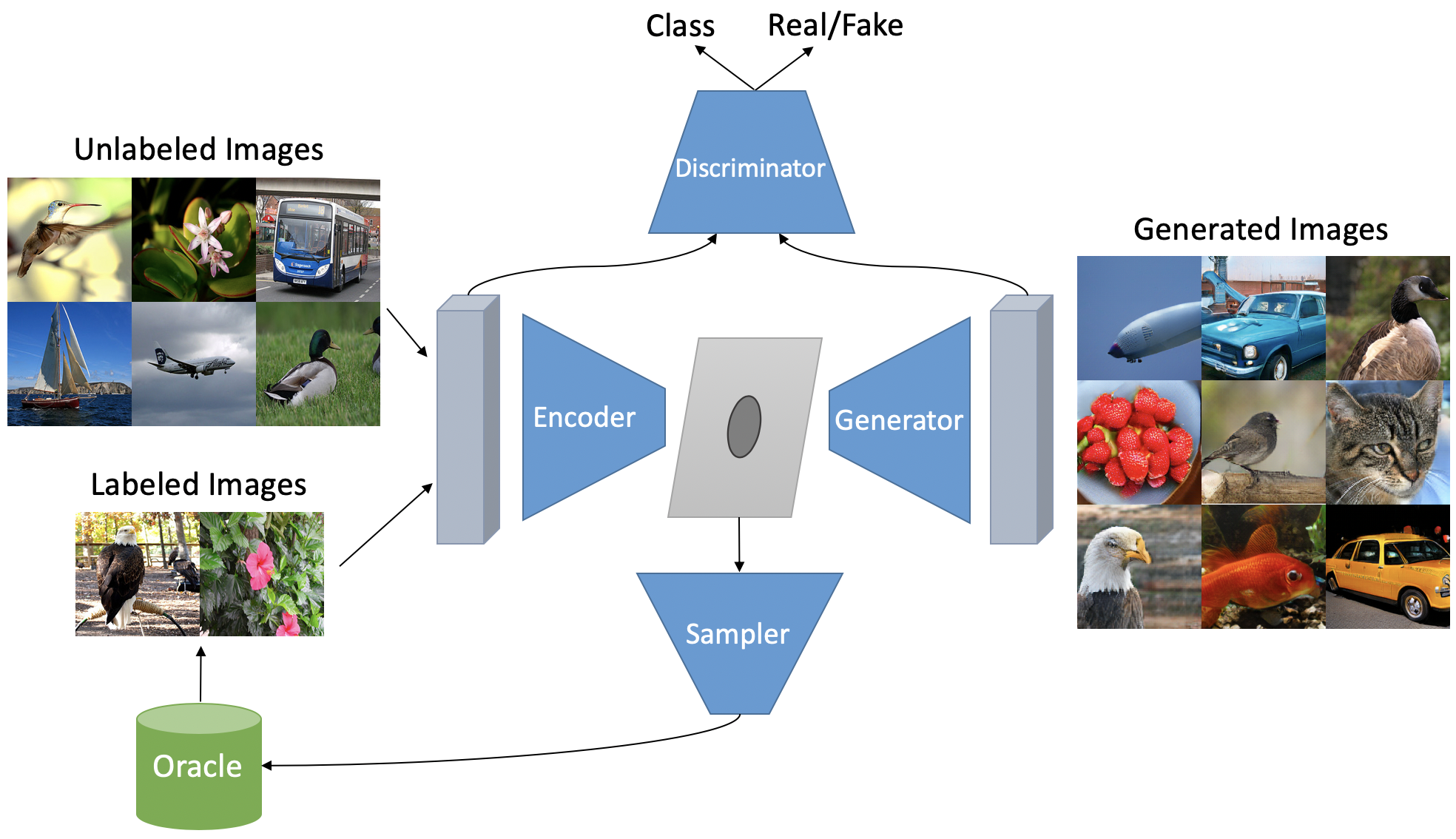}
  \vspace{-0.5cm}
\end{center}
   \caption{At each iteration, our model learns a latent representation of both labeled and unlabeled images and simultaneously trains an efficient classifier using labeled and class-conditional generated images. Then, it selects the most informative unlabeled images (based on their latent representation) to be labeled by the oracle for the next iteration.}
\label{fig:pull_figure}
\end{figure}

In this work, we introduce an active learning model based on the key observation that pool-based active learning approaches, which have access to a large unlabeled pool of data, can more effectively use this unlabeled data in the incremental training of the classifier itself. Particularly at small labeling budgets, this unlabeled data can provide valuable information about the underlying structure of the data, which is difficult to obtain from very few training examples, when directly integrated with the training of the classifier. We propose a model that utilizes a state-of-the-art variational adversarial acquisition function \cite{sinha2019variational} (which aims to select examples not well represented in the training set), but within a framework for efficiently training the classifier using both labeled and unlabeled data. Importantly, we perform the semi-supervised training by embedding the classifier within a GAN model that allows modeling the underlying structure of the unlabeled data to infer additional class labels for generated images that can be used to train the classifier. We share the encoder and decoder of the variational adversarial acquisition function as the encoder and generator of the conditional (bidirectional) GAN, and co-train the acquisition function and conditional GAN jointly. This allows, for example, the classifier in the GAN to improve the shared generator/decoder and hence the acquisition function. Vice versa, the improved acquisition function can then also better select training examples that most improve the classifier in the GAN.

We evaluate our proposed model on both standard image classification benchmarks for active learning---MNIST, SVHN, and CIFAR-10, as well as on more complex datasets---CelebA and ImageNet. Our model significantly outperforms prior active learning approaches on all these datasets, providing strong substantiation that active learning greatly benefits from more effective utilization of the unlabeled pool of data.

\section{Related Work}

\textbf{Active Learning}. Active learning has been widely studied and most of the early works can be found in the classical survey of \cite{settles2009active}. Current approaches can be categorized as query-acquiring (pool-based) and query-synthesizing algorithms. Pool-based methods use various acquisition functions for selecting the most informative examples, while query-synthesizing algorithms use generative models to generate informative examples. 

Pool-based approaches use several sampling strategies such as information-based  \cite{mackay1992information}, ensemble  \cite{mccallumzy1998employing} \cite{freund1997selective}, and uncertainty-based methods \cite{tong2001support} \cite{joshi2009multi} \cite{wang2016cost}. These methods are proposed in both Bayesian and non-Bayesian frameworks. In non-Bayesian classical active learning approaches, uncertainty heuristics such as distance from the decision boundary \cite{tong2001support} \cite{brinker2003incorporating}, highest entropy \cite{joshi2009multi}, and expected risk minimization \cite{tong2001support} have been widely investigated. Bayesian active learning methods use probabilistic models such as Gaussian processes \cite{kapoor2007active} or Bayesian neural networks \cite{ebrahimi2017gradient} to estimate uncertainty. \cite{houlsby2011bayesian} proposed the Bayesian active learning by disagreement (BALD) method in which the acquisition function is measured by the mutual information of the training examples with respect to the model parameters. \cite{gal2017deep} showed the relationship between uncertainty and dropout to estimate uncertainty in prediction in neural networks and applied it in active learning. \cite{sener2017active} proposed to use Core-set for selecting the subset of unlabeled images, where they minimize the Euclidean distance between the sampled points and the points that were not sampled in the feature space of the trained model. Also recently, \cite{kirsch2019batchbald}, \cite{ash2019deep}, and \cite{pinsler2019bayesian} proposed the batch active learning where they optimize for diversity as well as uncertainty.

Instead of querying the most informative instances from an unlabeled pool, \cite{zhu2017generative} introduced a generative adversarial active learning (GAAL) model to produce new synthetic examples that are informative for the current model. As they mainly rely on the generated samples for training the classifier, their performance highly depends on the quality and diversity of generated images. Also, the GAN model in \cite{zhu2017generative} is not fine-tuned as training progresses, therefore the generator and discriminator do not co-evolve. 

A few recent works also use generative models for active learning. \cite{tran2019bayesian} 
propose a Bayesian generative active deep learning approach where they combine active learning and data augmentation. However, they only use the labeled dataset for training their model and they use BALD \cite{gal2017deep} acquisition function for selecting new data points.

Also, \cite{sinha2019variational} suggests variational adversarial active learning, where they train a variational auto-encoder (VAE) and a discriminator to learn a latent representation on both labeled and unlabeled data. Then they use the output of the discriminator (between labeled and unlabeled images) as a measure of uncertainty for selecting from unlabeled data. In contrast to our work, they only use labeled images to train the classifier, and they use the VAE  only for acquiring unlabeled images. Contemporaneously with our own work, \cite{simeoni2019rethinking} also suggest using unlabeled data at model training, however, they do not show a significant improvement for using different sampling strategies as they perform active learning and semi-supervised learning separately. A key strength of our model is the idea that it learns the classifier and the acquisition function jointly using both labeled and unlabeled images.

\textbf{Generative Adversarial Networks}. Recent applications of GANs have shown that they can produce excellent samples. \cite{brock2018large} introduced BigGAN which uses class conditional generators trained on ImageNet to generate high-fidelity natural images. \cite{lucic2019high} later proposed S3GAN, where they use self- and semi-supervised learning methods to achieve the state-of-the-art sample quality in generating high resolution and diverse natural images. Their model matches the BigGAN in terms of quality of image generation using only $10\%$ of labels. \cite{donahue2016adversarial} and \cite{dumoulin2016adversarially} proposed an extension to the GAN model called bidirectional GAN (BiGAN) that augments the standard GAN with an encoder module mapping real data to the latent space, and the inverse of the mapping is learned by the generator. They showed that this model forms a good representation learner and can capture complex data distributions. Recently, \cite{donahue2019large} combined BiGAN and BigGAN to introduce BigBiGAN which achieves state-of-the-art representation learning on ImageNet. We leverage these ideas for learning a common representation for both labeled and unlabeled images; however, the main focus of our method is training the classifier and selecting the most informative examples. 

\section{Method}

\begin{figure*}[t]
\begin{center}
  \includegraphics[width=0.9\linewidth]{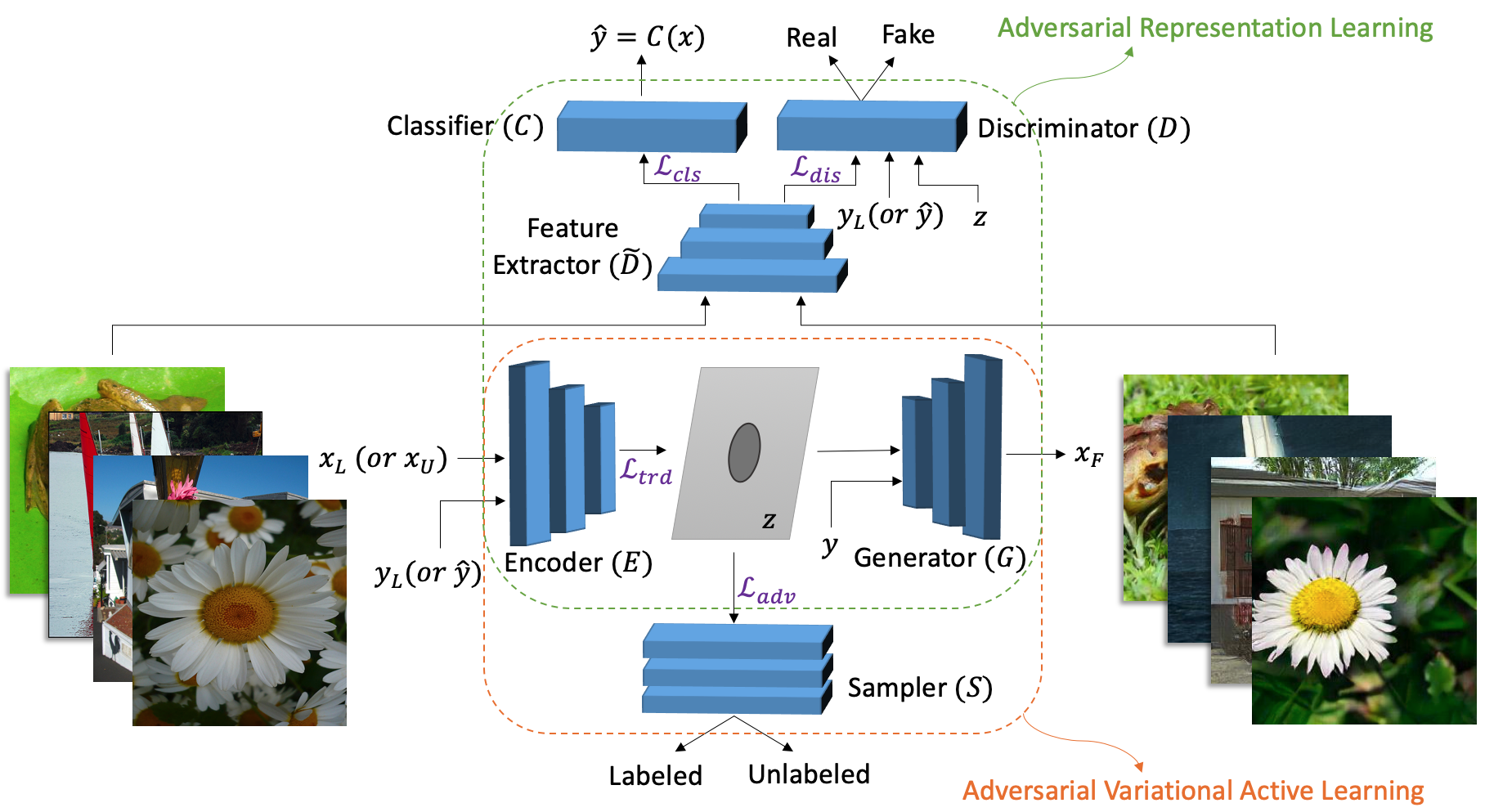}
\end{center}
   \caption{Our model consists of two main parts: 1) Adversarial variational active learning \cite{sinha2019variational} which attempts to learn a representation for labeled and unlabeled images in the latent space and select the most informative unlabeled images to be labeled next based on their latent code. 2) Adversarial representation learning \cite{donahue2019large} which couples the encoder-generator with a powerful discriminator and enhances the learned representation. We share the same features from the discriminator for training our classifier. We call the full model adversarial representation active learning. By co-training all parts of the model together, we can actively learn a classifier with much fewer labeled images as we will show in our experiments.}
\label{fig:model}
\end{figure*}

Our method addresses the standard active learning problem for image classification, where we have image space $\mathcal{X} \subseteq \mathbf{R}^{d_x}$, label space $\mathcal{Y} = \{1, 2, \dots, K\}$, and $K$ is the number of classes. Let $(x_L, y_L)$ denote a sample (image, label) pair belonging to labeled data $\mathcal{D}_L = (X_L, Y_L)$, and $x_U$ denote a sample image belonging to the pool of unlabeled data $\mathcal{D}_U = (X_U)$. Our goal is to train the most label-efficient classifier, i.e., with the highest classification accuracy for any given labeled dataset size $|\mathcal{D}_L|$. The active learner is allowed to iteratively select a fixed sampling budget $b$ number of samples from the unlabeled pool ($x_U \sim X_U$), to be annotated by an oracle and added to the labeled dataset $\mathcal{D}_L$ for the next iteration, using a sample acquisition function $S(x_U)$.

In a nutshell, our model introduces an approach for utilizing the unlabeled pool of data in addition to the active-labeled dataset for training the target classifier. The model uses a variational adversarial acquisition function as the sampling function. Our key contribution is the integration of this sampling function within a semi-supervised framework for training a classifier, which allows incorporation of the unlabeled data pool. Specifically, we use a semi-supervised conditional GAN, where the encoder and generator can be shared with the acquisition function and co-trained such that both the acquisition function and conditional GAN functions can mutually benefit from each others' improvement. Following, we first describe the variational adversarial acquisition function in Sec.~\ref{sec:vaal}. We then present our semi-supervised framework based on a conditional GAN for incorporating unlabeled data, in Sec.~\ref{sec:unlabeled-data}. Finally, in Sec.~\ref{sec:co-training}, we describe the co-training of our full model.

\subsection{Variational adversarial acquisition function}
\label{sec:vaal}
We use the variational adversarial active learning (VAAL) acquisition function~\cite{sinha2019variational} in our active learning method. This approach selects data examples for labeling that are not already well-represented in the labeled training set, by using a variational autoencoder with both reconstruction and adversarial losses. Specifically, the core of VAAL is a variational autoencoder consisting of encoder $E(x)$ which maps images to a latent representation, and decoder (i.e., generator) $G(z)$ which reconstructs images from latent representation $z$. A $\beta$-VAE reconstruction loss is used to perform transductive learning of a representation space from both labeled data $\mathcal{D}_L$ and unlabeled data $\mathcal{D}_U$:
\begin{align} \label{vaal_loss_trd}
    \mathcal{L}_{VAAL}^{trd} & = \mathbb{E} [p_{\theta_G}(x_L|z_L)] + \mathbb{E} [p_{\theta_G}(x_U|z_U)] \nonumber \\
    &- \beta D_{KL} (q_{\theta_E}(z_L|x_L)||p(z)) \nonumber \\
    &- \beta D_{KL} (q_{\theta_E}(z_U|x_U)||p(z)) \nonumber \\
    &= \mathcal{L}_{E-VAAL}^{trd} + \mathcal{L}_{G-VAAL}^{trd}
\end{align}

where $p_{\theta_G}$ and $q_{\theta_E}$ are the encoder and generator parameterized by $\theta_G$ and $\theta_E$, respectively. $\beta$ is the Lagrangian parameter for the optimization problem.

In addition to the VAE (i.e., the encoder-decoder), VAAL uses a discriminator which takes a latent representation $z$ as input and attempts to estimate the probability that the corresponding data comes from the labeled data. Once trained, this serves as the sampling function. We, therefore, denote the discriminator as S(z): if $S(z)$ is low, then the discriminator is very confident that the data point is unlabeled, and it is likely unrepresentative of the labeled set and a good candidate for labeling. The discriminator is trained together with the VAE in an adversarial manner, such that the VAE encoder tries to map the labeled and unlabeled data into the same latent space with a similar probability distribution, while the discriminator tries to distinguish labeled from unlabeled data. The encoder loss, therefore, has an additional term
\begin{align} \label{vaal_loss_adv}
    \mathcal{L}_{VAAL}^{adv} = &- \mathbb{E} [\log(S(q_{\theta_E}(z_L|x_L)))] \nonumber \\
    &- \mathbb{E} [\log(S(q_{\theta_E}(z_U|x_U)))]
\end{align}
such that the total encoder and generator loss is
\begin{align} \label{vaal_loss_e_g}
    \mathcal{L}_{E-VAAL} + \mathcal{L}_{G-VAAL} = & \lambda_{trd} \mathcal{L}_{VAAL}^{trd} + \lambda_{adv} \mathcal{L}_{E-VAAL}^{adv}
\end{align}
while the discriminator (sampler) loss is 
\begin{align} \label{vaal_loss_s}
    \mathcal{L}_{S-VAAL} = &- \mathbb{E} [\log(S(q_{\theta_E}(z_L|x_L)))] \nonumber \\
    &- \mathbb{E} [\log(1 - S(q_{\theta_E}(z_U|x_U))))]
\end{align}

\subsection{Semi-supervised framework for incorporating unlabeled data}
\label{sec:unlabeled-data}

Our framework for incorporating unlabeled data is based on the observation that the decoder in the VAAL acquisition function can be repurposed to additionally provide information about the unlabeled data and its underlying structure directly to a target classifier that we wish to train, by adapting it to simultaneously serve as the generator in a semi-supervised, class conditional GAN. The use of a class conditional GAN is key for several reasons. The first is that a class conditional GAN (as opposed to an unconditional GAN) contains a classifier component in the discriminator that can naturally serve as the target classifier for active learning. The second reason is that while the unconditional decoder in VAAL can only generate images from the unlabeled data distribution without any knowledge of classes, by adapting the decoder to simultaneously be the generator of a conditional GAN, it can generate class-conditional images (leveraging its exposure to the large unlabeled dataset) to improve training of the target classifier in the conditional GAN's discriminator. Therefore, the discriminator is decomposed into a learned discriminator representation $\tilde{D}$ which is fed into linear classifier $c_{r/f}$ for predicting real/fake and linear classifier $c_{cl}$ for predicting the class label. Using a similar approach as \cite{lucic2019high}, we also take into account the encoded latent variable $z$ and the real class label or inferred class label in linear classifier $c_{r/f}$ for better predicting real/fake images. We denote $D(x, z, y) = c_{r/f}(\tilde{D}(x), z, y)$ for real/fake predictions and $C(x) = c_{cl}((\tilde{D}(x)))$ for class label predictions. All the modules in our model (Generator, Discriminator, Encoder, and Sampler) are depicted in Fig. \ref{fig:model}.

To adapt the VAAL acquisition function for our semi-supervised framework, we make the encoder and generator (decoder) class-conditional. The encoder, generator, and sampler (discriminator) losses for the acquisition function that we use (Eq. \ref{vaal_loss_e_g} and Eq. \ref{vaal_loss_s}) therefore become:
\begin{align} 
    \mathcal{L}_{E}^{acq} &+ \mathcal{L}_{G}^{acq} = \lambda_{trd} \mathbb{E} [p_{\theta_G}(x_L|z_L, y_L)] \nonumber \\
    &+ \lambda_{trd}\mathbb{E} [p_{\theta_G}(x_U|z_U, C(x_U))] \nonumber \\
    &- \lambda_{trd} \beta D_{KL} (q_{\theta_E}(z_L|x_L, y_L)||p(z))) \nonumber \\
    &- \lambda_{trd} \beta D_{KL} (q_{\theta_E}(z_U|x_U, C(x_U))||p(z))) \nonumber \\
    &- \lambda_{adv} \mathbb{E}[\log(S(q_{\theta_E}(z_L|x_L, y_L)))] \nonumber \\
    &- \lambda_{adv} \mathbb{E}[\log(S(q_{\theta_E}(z_U|x_U, C(x_U))))] \label{e_g_loss_acq} \\
    \nonumber \\
    \mathcal{L}_{S}^{acq} = &- \mathbb{E} [\log(S(q_{\theta_E}(z_L|x_L, y_L)))] \nonumber \\
    &- \mathbb{E} [\log(1 - S(q_{\theta_E}(z_U|x_U, C(x_U))))] \label{s_loss_acq}
\end{align}

Next, we describe the generator and decoder of our semi-supervised, conditional GAN framework. We also add an encoder as in BiGAN~\cite{donahue2016adversarial}, which has been shown to improve classification performance for more complex data \cite{donahue2019large} and can also be shared with the acquisition function.

\textbf{Generator}. The objective of the generator in the conditional GAN framework is to generate class-conditional images that can fool the discriminator into predicting them as real images. Since it learns to generate by playing a min-max game with the discriminator (and therefore learns from both labeled and unlabeled data), the generated images convey information about the structure in the unlabeled data that augments the labeled data when training the target classifier in the discriminator (described in the next section). The generator loss is the standard loss for conditional GANs:

\vspace{-0.5cm}
\begin{align} \label{g_loss_gan}
    \mathcal{L}_{G}^{gan} &= \mathbb{E}_{z \sim p(z), y \sim p(y)} [\log(1 - D(G(z, y), z, y))]
\end{align}

\textbf{Discriminator}. The discriminator (Fig. \ref{fig:model}) structure also follows that of standard conditional GANs, containing both a real/fake discriminator network and a classification network, however, we use it in the semi-supervised setting. Importantly, the classification network serves as the target classifier for which we will try to maximize accuracy within the active learning problem. The loss for the real/fake discriminator network is:
\begin{align} \label{d_loss_dis}
    \mathcal{L}_{D}^{gan} = &- \mathbb{E}_{(x, y) \sim (X_L, Y_L)} [\log(D(x, E(x, y), y))] \nonumber \\
    &- \mathbb{E}_{x \sim X_U} [\log(D(x, E(x, C(x)), C(x)))] \nonumber \\
    &+ \mathbb{E}_{z \sim p(z), y \sim p(y)} [\log(D(G(z, y), z, y))]
\end{align}
where the first term corresponds to the standard discriminator loss for labeled data and the third term corresponds to the standard loss for generated data. The second term corresponds to the discriminator loss for unlabeled data, where the labels for these examples are inferred through the classifier $C(x)$ in the descriminator, described next. Note that the real/fake discriminator is a function of the data $x$ or $G(z,y)$, the class $y$ or $C(x)$ due to being a conditional GAN, as well as the latent representation $E(x,y)$, $E(x,C(x))$, or $z$ due to the BiGAN structure with a concurrently learned encoder.

The loss for the classification network in the discriminator is:
\begin{align} \label{d_loss_cls}
    \mathcal{L}_{D}^{cls} = &- \mathbb{E}_{(x, y) \sim (X_L, Y_L)} [\log p(C(x) = y)] \nonumber \\
    &- \mathbb{E}_{z \sim P(z), y \sim P(y)} [\log p(C(G(z, y)) = y)]
\end{align}
Here the first term is the cross-entropy loss for real images that have ground truth labels, and the second term is the cross-entropy loss for generated images that have corresponding labels that were used for the generation. This second term allows the classifier to benefit from the unlabeled data used to train the generator. Since the classification network and real/fake discriminator network have a shared trunk, the real/fake supervision additionally enables learning a stronger shared feature representation that can further improve classification performance.

\textbf{Encoder}. In addition to the generator and discriminator, we add an encoder (shown in Fig. \ref{fig:model}) in our conditional GAN following BiGAN~\cite{donahue2016adversarial}. This has been shown to improve classification performance for more complex data \cite{donahue2019large} and is a natural choice since our acquisition function already has an encoder that can be shared. Our encoder loss following BiGAN is:
\begin{align} \label{e_loss_gan}
    \mathcal{L}_{E}^{gan} = &- \mathbb{E}_{(x, y) \sim (X_L, Y_L)} [\log(D(x, E(x, y), y))] \nonumber \\
    &- \mathbb{E}_{x \sim X_U} [\log(D(x, E(x, C(x)), C(x)))]
\end{align}
where the first term corresponds to the discriminator loss for labeled images and the second term correspond to unlabeled images.

\begin{algorithm}[t]
    \caption{Adversarial Representation Active Learning} \label{alg}
    \begin{algorithmic}[1]
        \renewcommand{\algorithmicrequire}{\textbf{Input:}}
        \renewcommand{\algorithmicensure}{\textbf{Output:}}
        \REQUIRE Labeled pool $\mathcal{D}_L$, Unlabeled pool $\mathcal{D}_U$, Labeling budget, Initialized models for $\theta_G$, $\theta_E$, $\theta_D$, and $\theta_S$
        \REPEAT 
        \STATE Pick samples $X_s$ from $\mathcal{D}_U$ with $\min_b\{S(E(X_s))\}$
        \STATE Label the selected samples by oracle, which forms $(X_s, Y_o)$
        \STATE Update labeled and unlabeled datasets:
        \STATE $\mathcal{D}_L \gets \mathcal{D}_L \cup (X_s, Y_o)$
        \STATE $\mathcal{D}_U \gets \mathcal{D}_U - X_s$
        \FOR{$e = 1$ \text{to epochs}}
        \STATE Sample $(x_L, y_L) \sim \mathcal{D}_L$
        \STATE Sample $x_U \sim \mathcal{D}_U$
        \STATE Compute $\mathcal{L}_E + \mathcal{L}_G$ using Eq. \ref{e_g_loss}
        \STATE Update $E$ and $G$ by descending stochastic gradient
        \vspace{-10pt}
        \STATE $\theta'_E \gets \theta_E - \alpha_E \nabla (\mathcal{L}_E + \mathcal{L}_G)$
        \STATE $\theta'_G \gets \theta_G - \alpha_E \nabla (\mathcal{L}_E + \mathcal{L}_G)$
        \STATE Compute $\mathcal{L}_D$ using Eq. \ref{d_loss}
        \STATE Update D by descending stochastic gradient
        \vspace{1pt}
        \STATE $\theta'_D \gets \theta_D - \alpha_D \nabla \mathcal{L}_D$
        \STATE Compute $\mathcal{L}_S$ using Eq. \ref{s_loss}
        \STATE Update S by descending stochastic gradient
        \vspace{1pt}
        \STATE $\theta'_S \gets \theta_S - \alpha_S \nabla \mathcal{L}_S$
        \ENDFOR
        \UNTIL labeling budget is finished
        \RETURN Trained $\theta_E$, $\theta_G$, $\theta_D$, and $\theta_S$
    \end{algorithmic} 
\end{algorithm}

\subsection{Co-training of full model}
\label{sec:co-training}
To perform active learning, the acquisition function and the conditional GAN presented above are jointly co-trained, where the losses for the full model are:
\begin{align} 
    \vspace{-1cm}
    \mathcal{L}_{E} + \mathcal{L}_{G} &= \mathcal{L}_{E}^{acq} + \mathcal{L}_{G}^{acq} \nonumber \\
    &+ \lambda_{gan} \mathcal{L}_{E}^{gan} + \lambda_{gan} \mathcal{L}_{G}^{gan} \label{e_g_loss} \\
    \mathcal{L}_{D} &= \lambda_{gan} \mathcal{L}_{D}^{gan} + \lambda_{cls} \mathcal{L}_{D}^{cls} \label{d_loss} \\
    \mathcal{L}_{S} &= \mathcal{L}_{S}^{acq} \label{s_loss}
\end{align}

Note that the generator and encoder are shared between the acquisition function and the conditional GAN, while the discriminator and the sampler are not. After every selection and labeling of new samples, all components of the model are updated using the new labeled dataset $D_L$ and unlabeled pool $D_U$. The full algorithm is presented in Alg.~\ref{alg}.

\section{Experiments}

We evaluate the performance of our model on a wide range of datasets and compare it to prior state-of-the-art active learning methods. We assess performance by measuring the accuracy of the classifier trained during the active learning procedure versus the number of labeled images used in the training.

\textbf{Baselines}: We compare our results for actively training our classifier against the following baselines: 1) \textbf{Uncertainty-based methods}: In these approaches, unlabeled images will be labeled based on the uncertainty in the classifier's prediction for them. We perform active learning using Max-Entropy \cite{shannon1948mathematical}, Variation Ratios \cite{freeman1965elementary}, and Mean STD \cite{kendall2015bayesian} among the methods in this class. We observed that these achieve similar performance; thus, we only compare against the Max-Entropy method. 2) \textbf{Bayesian methods}: In Bayesian frameworks, probabilistic models such as Gaussian processes and Bayesian neural networks are used to estimate the expected improvement by each query. \cite{gal2017deep} used dropout as an approximation to Bayesian inference and used Bayesian Active Learning by Disagreement (BALD) for Bayesian active learning on image data. We report the performance of this method in our experiments.
Another recent approach in this class is Bayesian Generative Active Deep Learning (BGADL) \cite{tran2019bayesian}. As the authors reported, their model does not converge for very few numbers of labels (which is the setting in our experiments), so we do not report the results for this method. 3) \textbf{Variational adversarial active learning}: We also compare our model with the recent state-of-the-art method Variational Adversarial Active Learning (VAAL) \cite{sinha2019variational}. We use a sampling strategy similar to theirs in our method; however, VAAL trains the classifier separately only on the selected labeled images. 4)  \textbf{Random}: We show results using random sampling, in which samples are uniformly sampled from the unlabeled pool, and the classifier is then trained on the labeled data. 5) \textbf{Full training of our model}: As another baseline, we compare the performance of the model during the learning iterations with the fully trained model (when we use all the labels.) This serves as an upper bound for our performance and shows how fast our method converges to the best accuracy.


Following, we first evaluate our model versus the baselines on classic active learning benchmarks (MNIST, SVHN, CIFAR-10) in Sec. \ref{sec:exp_classic}, and then on more complex datasets (CelebA, ImageNet) in Sec. \ref{sec:exp_complex}. Finally, in Sec. \ref{sec:ablation}, we perform ablation studies on our model.


\subsection{Performance on classic active learning benchmarks} \label{sec:exp_classic}

\begin{figure}[h]
\begin{center}
  \includegraphics[width=1\linewidth]{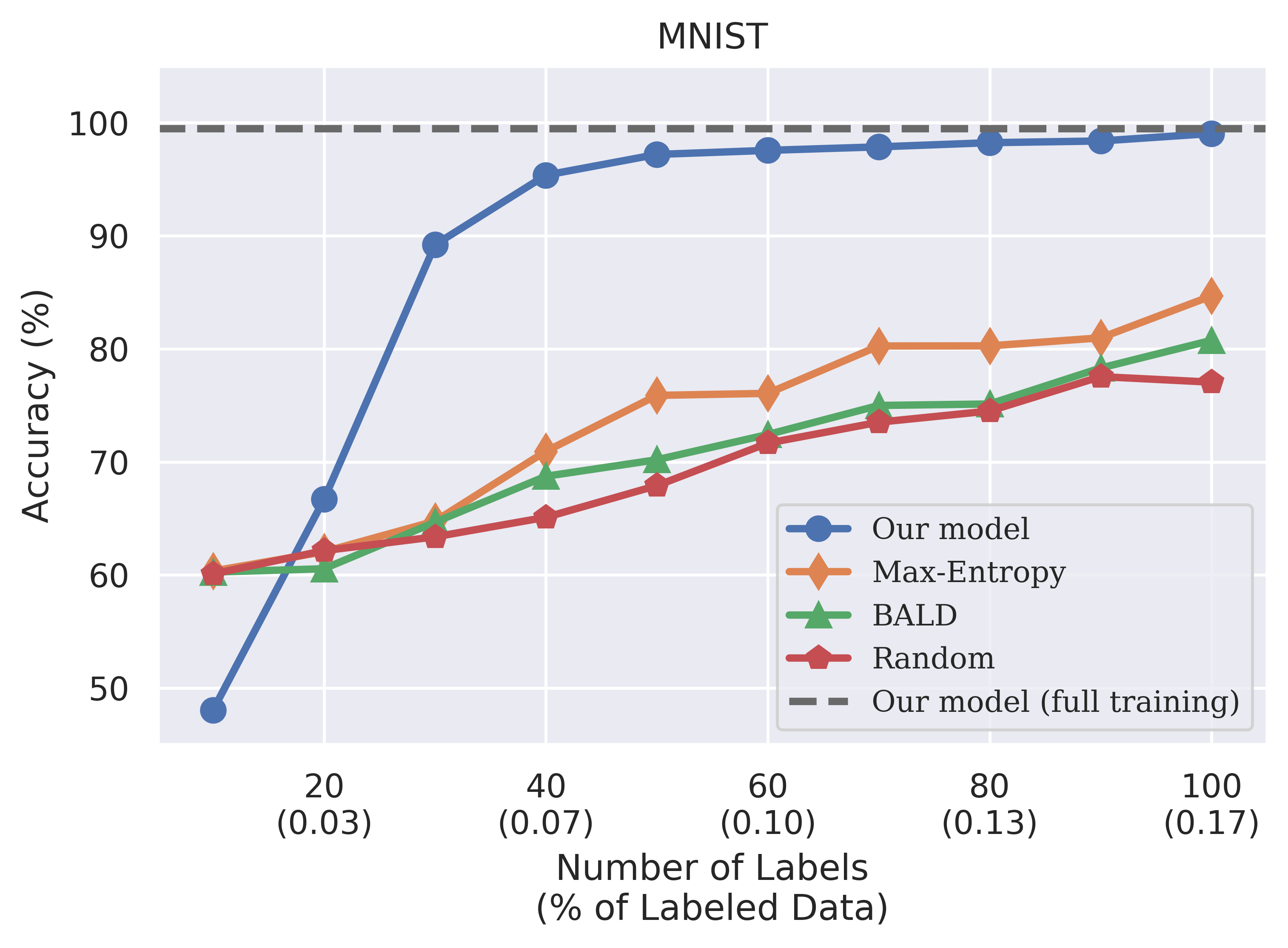}
  \includegraphics[width=1\linewidth]{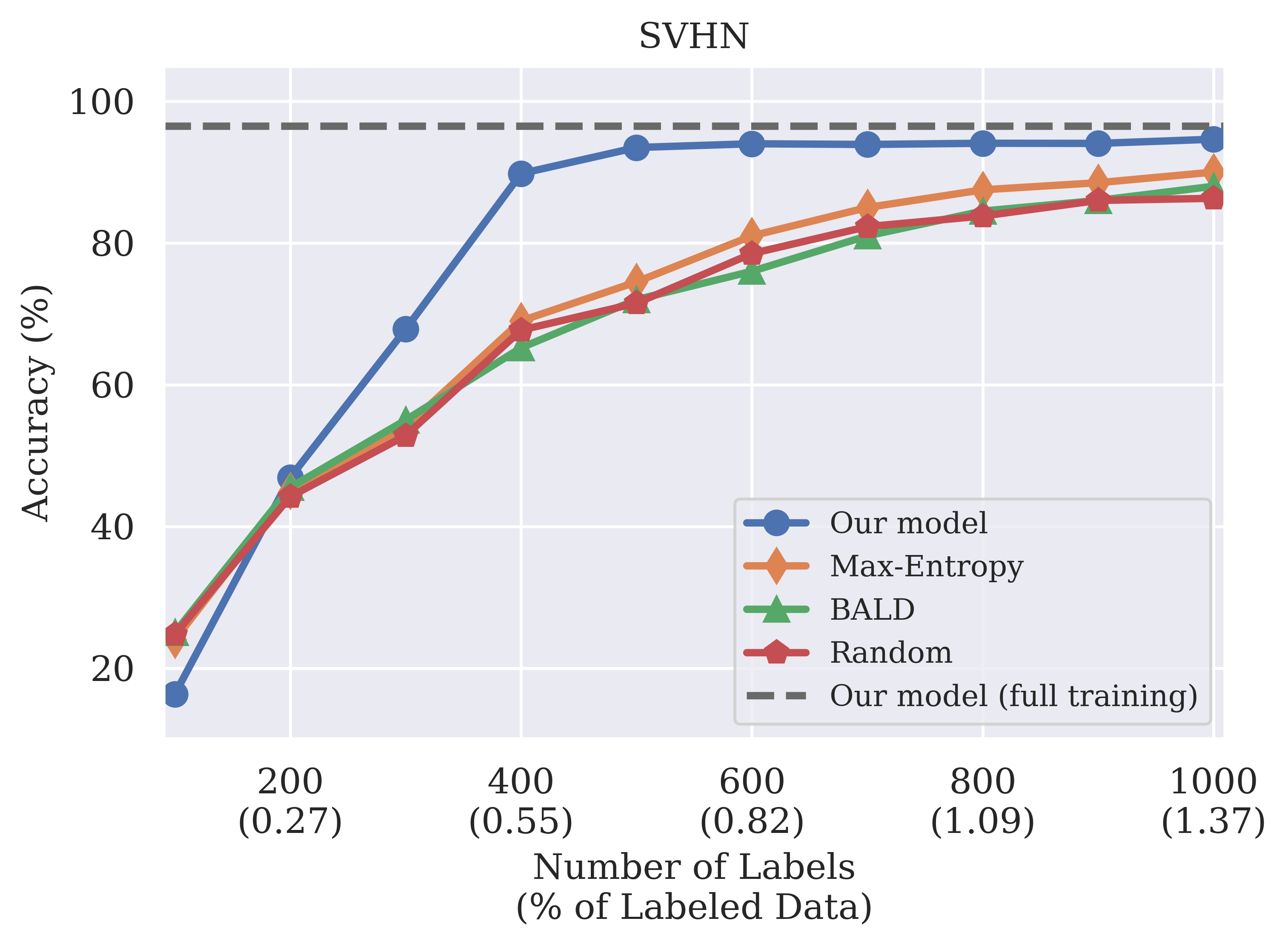}
  \includegraphics[width=1\linewidth]{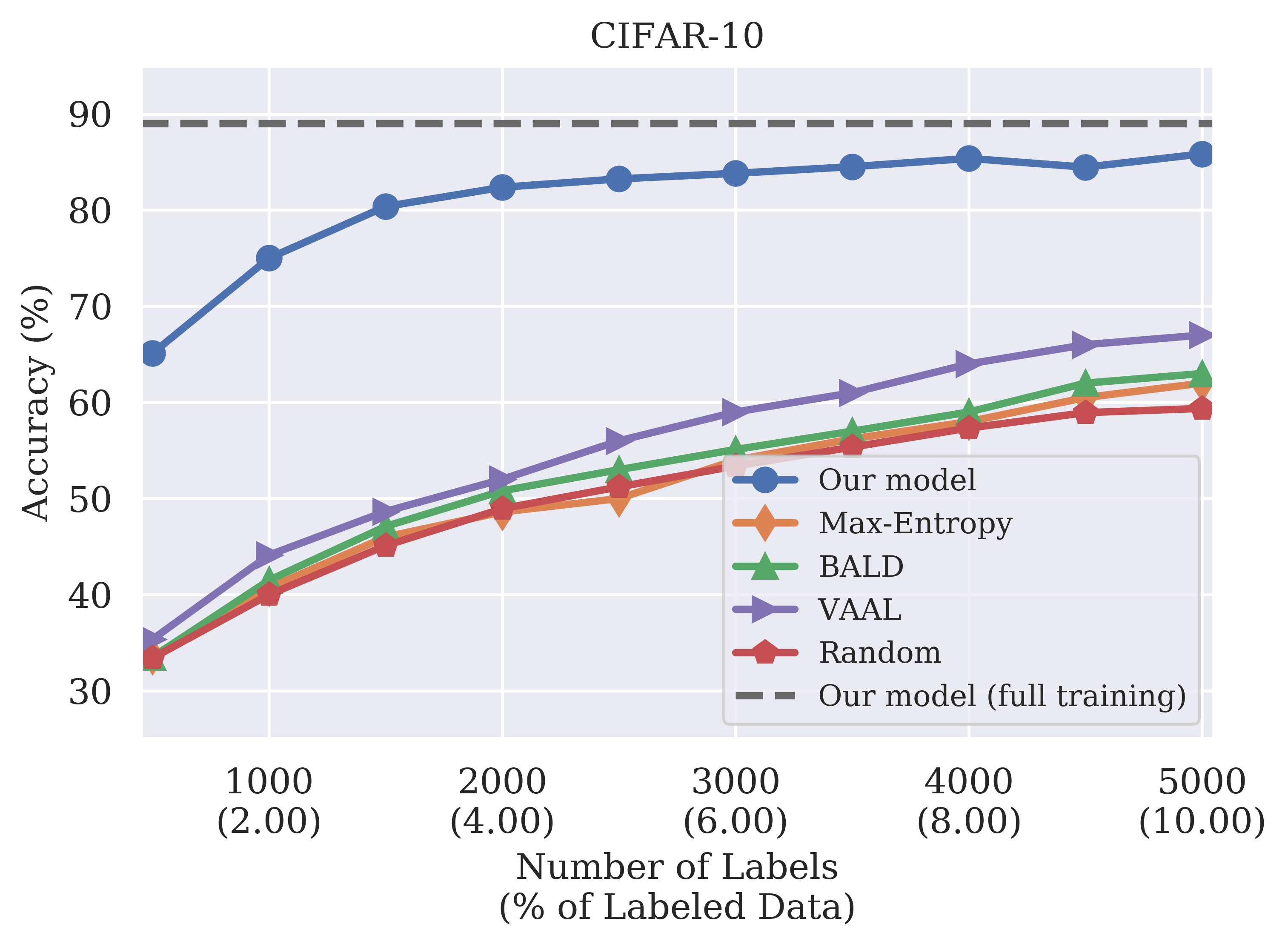}
\end{center}
   \vspace{-0.5cm}
   \caption{Our model performance compared to Max-Entropy \cite{shannon1948mathematical}, BALD \cite{gal2017deep}, VAAL \cite{sinha2019variational}, Random baseline, and full training on MNIST, SVHN, and CIFAR-10 datasets.}
   \vspace{-0.5cm}
\label{fig:exp_classic}
\end{figure}

\textbf{Datasets}: The MNIST \cite{lecun1998gradient} dataset contains $28 \times 28$ images of handwritten digits of 10 classes (with $60k$ training samples and $10k$ test samples). The SVHN \cite{netzer2011reading} and CIFAR-10 \cite{krizhevsky2009learning} datasets have images of size $32 \times 32$ of 10 classes (with $73257$ and $50k$ training samples, and $26032$ and $10k$ test samples, respectively). 

\textbf{Architecture and hyper-parameters}: For the MNIST dataset, we used a 3-layer MLP for both the encoder and generator and a 5-layer MLP with added Gaussian noise (with a standard deviation of $0.5$) between the layers for the discriminator. Our sampler module for this dataset is also a 5-layer MLP. Adam \cite{kingma2014adam} with a learning rate of $1 \times 10^{-3}$ is chosen as the optimizer for the encoder and generator modules, and with a learning rate of $3 \times 10^{-3}$ for the discriminator and sampler modules. For the SVHN and CIFAR-10 datasets, we use a CNN with 3 hidden layers for both the encoder and generator and 3 convolutional blocks, each with 3 layers and dropout (with a rate of $0.5$) between the blocks, for the discriminator. Our sampler module for this dataset is the same 5-layer MLP. Adam with a learning rate of $3 \times 10^{-4}$ is chosen as the optimizer for the encoder and generator modules and with a learning rate of $6 \times 10^{-4}$ for the discriminator and sampler modules. We also use $\lambda_{gan}=\lambda_{cls}=1, \lambda_{trd}=\lambda_{adv}=0.001,$ and $\beta=1$ as our hyperparameters. For these datasets we use a latent space with $d_z = 100$ dimensions. We also use the feature matching technique which is proposed in \cite{salimans2016improved} for matching the features in the discriminator for generated and unlabeled images. In this way, generated images will be a better representative of the unlabeled parts of the dataset.

\textbf{Implementation details}: 
We begin our experiments with an initial labeled pool with only $10$ labels for the MNIST dataset, $100$ labels for SVHN, and $500$ labels for CIFAR-10. We add the same number of labels at each iteration of active learning and report results for the first $10$ iterations. We also report the accuracy after full training of our model. 

\textbf{Our model performance}: Fig. \ref{fig:exp_classic} shows the performance of our model compared to the baselines on MNIST, SVHN, and CIFAR-10 datasets. Our model significantly outperforms all the baselines as it uses the unlabeled images as well as generated images in the training procedure. The performance gap is more clear especially for very few numbers of labels when unlabeled and generated images are more useful in the training. The performance of our model saturates quickly on the classic benchmarks as it achieves accuracy close to the full training accuracy with the size of the labeled dataset on the order of $0.1\%$ of labels for MNIST, $1\%$ for SVHN, and $10\%$ for CIFAR-10. By looking at the number of labels required to reach a specific accuracy, for instance, $65\%$ on CIFAR-10 dataset, our model only needs $500$ images to be labeled while this number is approximately $8000$ for VAAL. This shows the importance of using unlabeled data when training the model in an active manner, which ultimately results in more label efficient learning.

\subsection{Performance on more complex datasets} \label{sec:exp_complex}

\textbf{Datasets}: The large-scale CelebFaces Attributes (CelebA) \cite{liu2018large} dataset is a more challenging dataset with more than $200k$ celebrity images, each annotated with $40$ attributes. We split the dataset into $150k$ images for training and $50k$ images for testing. Then we resize all the images into $64 \times 64$ pixels and use max normalization in the prepossessing phase. Our task is to classify the images into two classes (male/female) based on the gender attribute annotation in the dataset. ImageNet \cite{deng2009imagenet} is a large dataset with more than $1.2M$ images of $1000$ classes. The validation set for this dataset contains $50k$ images. We augment our dataset by horizontally flipping the images. Then we resize all the images into $224 \times 224$ pixels and normalize them before feeding them into the model.

\begin{figure}[t]
\begin{center}
  \includegraphics[width=1\linewidth]{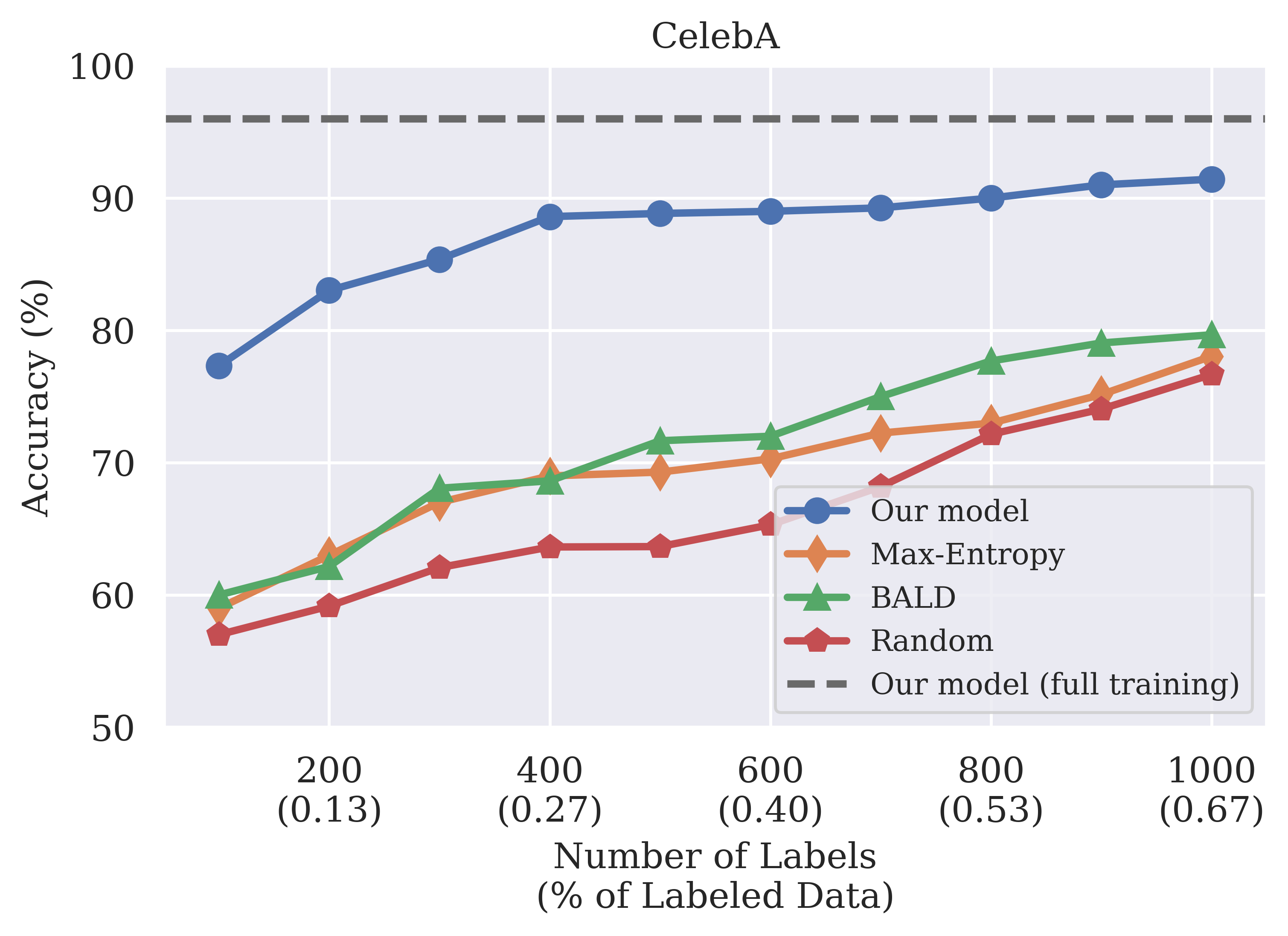}
  \includegraphics[width=1\linewidth]{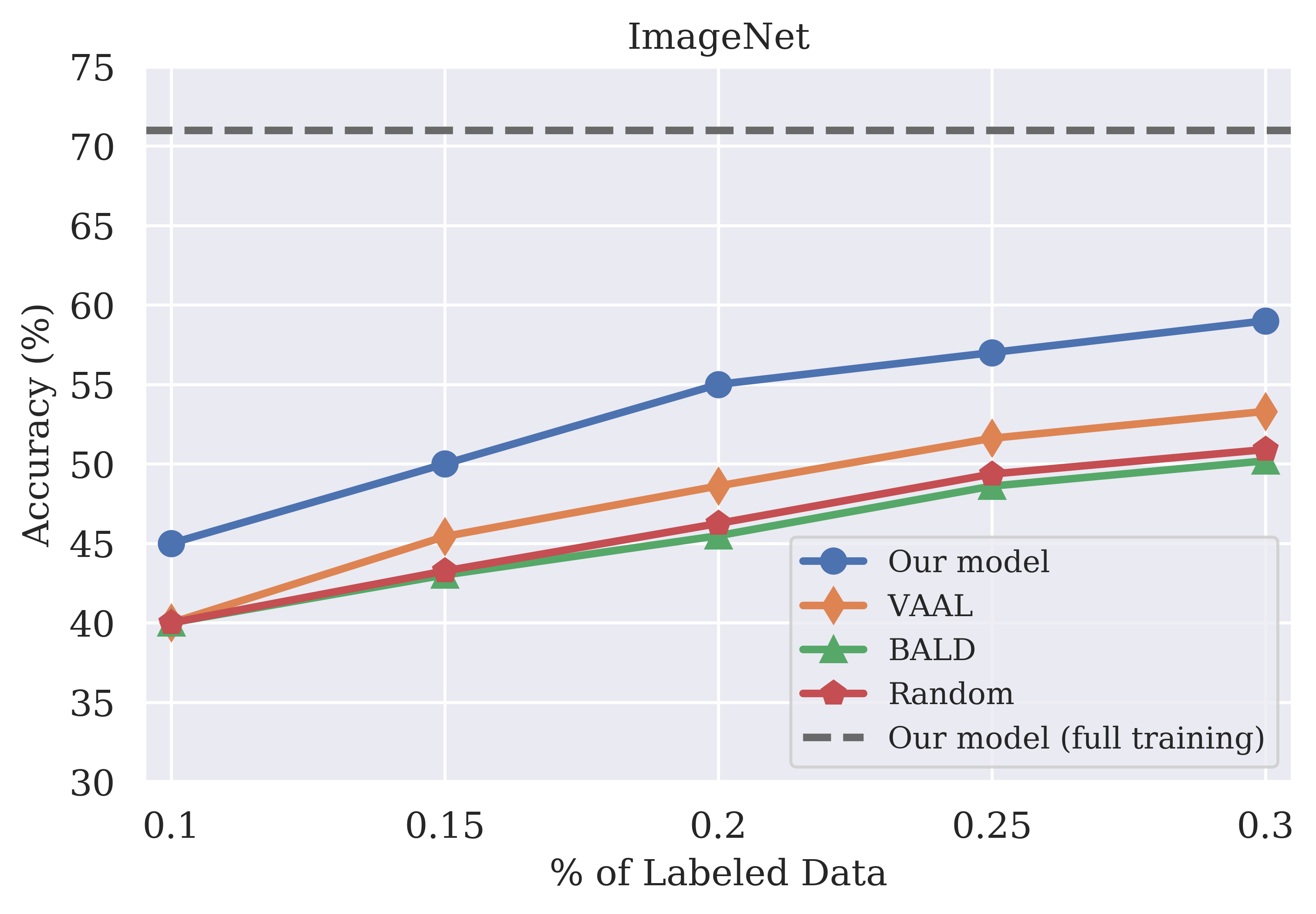}
\end{center}
   \vspace{-0.5cm}
   \caption{Our model performance compared to Max-Entropy \cite{shannon1948mathematical}, BALD \cite{gal2017deep}, VAAL \cite{sinha2019variational}, and Random baseline, and full training on CelebA and ImageNet datasets.}
   \vspace{-0.5cm}
\label{fig:exp_complex}
\end{figure}

\textbf{Architecture and hyper-parameters}: For the CelebA dataset, we used 4 residual blocks each with two convolutional layers for the encoder and generator, and another 4 residual blocks with added dropout layer (with a rate of $0.5$) for the discriminator. Adam with a learning rate of $1 \times 10^{-3}$ is chosen as the optimizer for all parts. For ImageNet, we use the same architecture and hyper-parameters as S3GAN in \cite{lucic2019high}, which is the state-of-the-art in image generation on the ImageNet dataset. The rest of hyper-parameters are the same as the previous experiments except for the latent space which has $d_z = 64$ dimensions. 

\textbf{Implementation details}: 
For the CelebA dataset, we begin our experiments with an initial labeled pool of only $100$ labels. We add the same number of labels for $10$ iterations. 
For experiments on ImageNet, due to our limitation in accessing a huge computation power, we trained a relatively smaller (compared to BigGAN \cite{brock2018large}) generator-encoder module as well as a sampler using a similar approach as \cite{sinha2019variational}, to select the most informative examples in each iteration by mapping both labeled and unlabeled data into the latent space. Then we added the high-quality fake images generated by a pretrained $S^3GAN$ model (with $256 \times 256$ pixels resolution) \cite{lucic2019high} on $10\%$ of labels to augment our labeled data, and finally, train the classifier network ($VGG16$) using the labeled and generated images. Therefore, our experiments here utilize a variant of our model where we are training the encoder and sampler modules (for selecting the most informative examples) separately from the generator and discriminator (for generating high-quality fake images and training the classifier on labeled and generated images).

\textbf{Our model performance}: As can be observed from Fig. \ref{fig:exp_complex}, our model significantly outperforms all the baselines on these datasets as well. Similar to what we observed on the previous datasets, on CelebA there is only a small gap between many active learning methods and the Random baselines. However, the performance gap between our model and the other baselines is significant and steady over the entire training procedure. This shows that our generative model approach can mitigate the lack of real labeled images. Fig. \ref{fig:exp_complex} also shows the performance of our model on the ImageNet dataset compared to other methods. As we are not co-training all the parts together, the performance gap from the model trained on all of the labels is bigger compared to previous datasets, however, it still significantly outperforms all the baselines.

\subsection{Ablation study} \label{sec:ablation}
We perform an ablation study on the CIFAR-10 dataset. For showing the effectiveness of each part of our model, we consider the following variants of ablation and compare the performances: 1) \textbf{No active learning}: we remove the sampler and adversarial loss for the encoder and use random sampling at each iteration of training the model. 2) \textbf{No encoder}: we only have the generator and discriminator modules and we use $BALD$ as our sampling strategy. 3) \textbf{No co-training}: similar to our experiment on ImageNet, we again use VAAL as our acquisition function. However, instead of utilizing the unlabeled data via co-training the encoder-generator pair with the discriminator, we add generated images from a pretrained $S^3GAN$ model to train the classifier. 4) \textbf{Random}: samples are uniformly sampled from the unlabeled pool and the classifier is trained on the labeled data.

\begin{figure}[t]
\begin{center}
  \includegraphics[width=\linewidth]{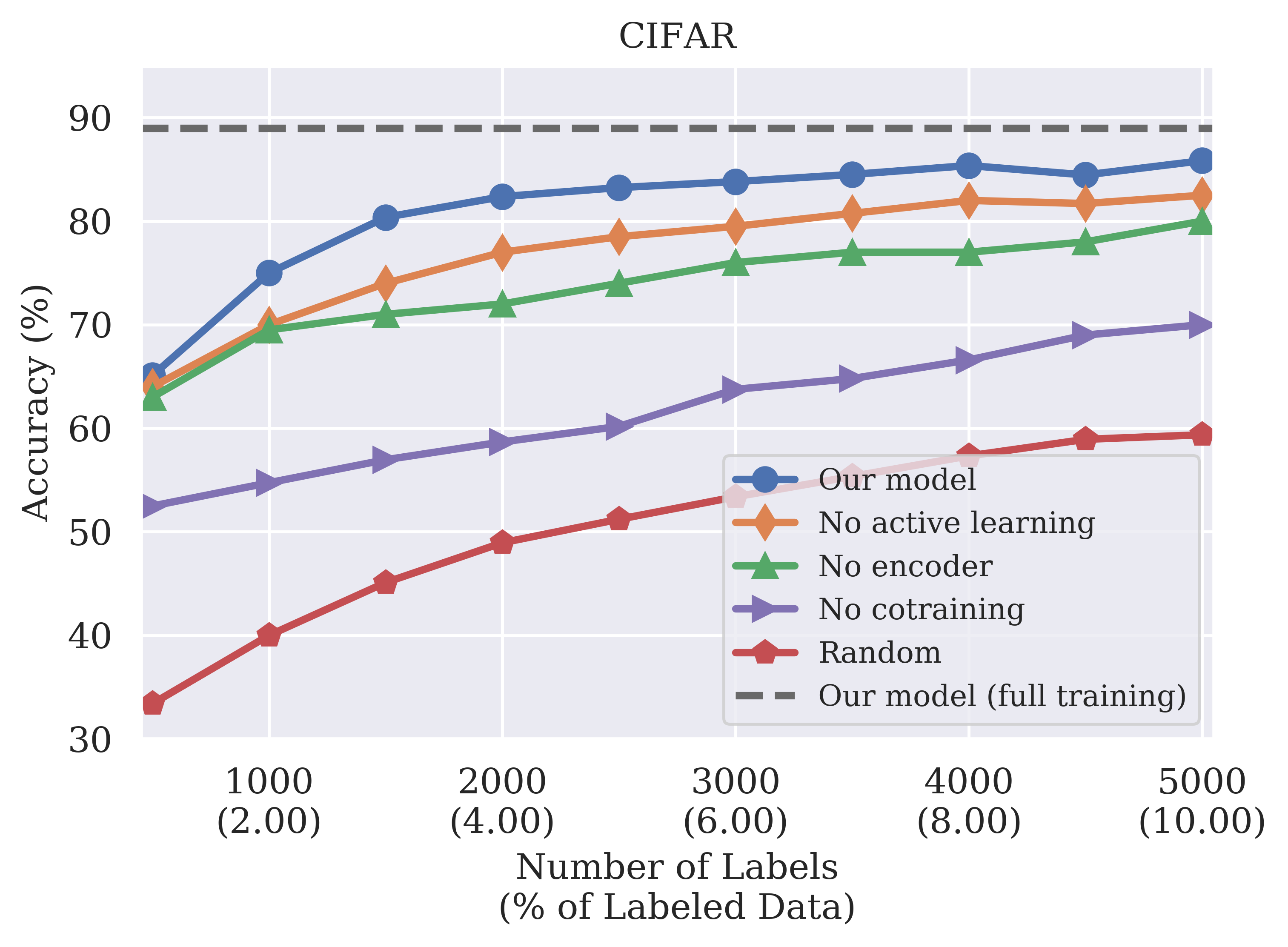}
\end{center}
   \vspace{-0.5cm}
   \caption{Ablation studies for analysing the effect of each component in our model on CIFAR-10 dataset.}
   \vspace{-0.5cm}
\label{fig:cifar_ab}
\end{figure}

As shown in Fig. \ref{fig:cifar_ab}, each module contributes to the final performance of the model. In the first variant, we use random sampling as our sampling strategy, therefore, it can be seen as the performance for using unlabeled images in addition to labeled images in our adversarial learning method without using active learning algorithms. Although our purpose in this work is not training semi-supervised generative models, we are also outperforming related semi-supervised learning with generative models approaches such as \cite{salimans2016improved}, which achieves $81.37\%$ accuracy using $4000$ labels. Our second variant captures the effect of the encoder in training our model.
In this setting, the model utilizes the generator and discriminator modules, but it does not perform the representation learning with labeled and unlabeled images (and therefore uses BALD instead of VAAL as the acquisition function), and cannot perform as well as our model. We also conduct an experiment similar to what we performed for the ImageNet dataset to better understand the effect of co-training all parts of the model together. Although we are using the same sampling strategy (VAAL) here and adding the class-conditional generated images in order to utilize the unlabeled data, its performance is still significantly below our model in which all parts are co-trained with each other. Finally, we have the Random baseline which is the fundamental baseline in active learning literature.

\section{Conclusion}
In this work, we proposed a new active learning method using deep generative models that takes advantage of both labeled and unlabeled images, for learning a representation that is used not only for selecting the most informative examples but also for utilizing unlabeled data in training the classifier. We demonstrated that our model significantly outperforms the previous state-of-the-art on a wide range of datasets (MNIST, CIFAR-10, SVHN, CelebA, ImageNet).

{\small
\bibliographystyle{ieee_fullname}
\bibliography{egbib}
}

\end{document}